\documentclass[10pt,twocolumn,letterpaper]{article}

\usepackage{cvpr}              
\usepackage{algorithm}
\usepackage{algpseudocode}

\usepackage[dvipsnames]{xcolor}

\definecolor{cvprblue}{rgb}{0.21,0.49,0.74}
\usepackage[pagebackref,breaklinks,colorlinks,citecolor=cvprblue]{hyperref}

\def\paperID{9} 
\def\confName{CVPR}
\def\confYear{2024}

\title{Segment Anything Model for Road Network Graph Extraction}

\author{
Congrui Hetang\\
Carnegie Mellon University\\
{\tt\small congruihetang@gmail.com}
\and
Haoru Xue \\
Carnegie Mellon University\\
{\tt\small haorux@andrew.cmu.edu}
\and
Cindy Le\\
Columbia University\\
{\tt\small xl2738@columbia.edu}
\and
Tianwei Yue\\
Carnegie Mellon University\\
{\tt\small tyue@alumni.cmu.edu}
\and
Wenping Wang\\
Carnegie Mellon University\\
{\tt\small wenpingw@alumni.cmu.edu}
\and
Yihui He\\
Carnegie Mellon University\\
{\tt\small he2@alumni.cmu.edu}
}

\begin{document}
\maketitle
\begin{abstract}
We propose SAM-Road, an adaptation of the Segment Anything Model (SAM) \cite{kirillov2023segment} for extracting large-scale, vectorized road network graphs from satellite imagery. To predict graph geometry, we formulate it as a dense semantic segmentation task, leveraging the inherent strengths of SAM. The image encoder of SAM is fine-tuned to produce probability masks for roads and intersections, from which the graph vertices are extracted via simple non-maximum suppression. To predict graph topology, we designed a lightweight transformer-based graph neural network, which leverages the SAM image embeddings to estimate the edge existence probabilities between vertices. Our approach directly predicts the graph vertices and edges for large regions without expensive and complex post-processing heuristics and is capable of building complete road network graphs spanning multiple square kilometers in a matter of seconds. With its simple, straightforward, and minimalist design, SAM-Road achieves comparable accuracy with the state-of-the-art method RNGDet++\cite{xu2023rngdetplus}, while being 40 times faster on the City-scale dataset. We thus demonstrate the power of a foundational vision model when applied to a graph learning task. The code is available at \url{https://github.com/htcr/sam_road}.
\end{abstract}    
\section{Introduction}
\label{sec:intro}

\begin{figure}
  \centering
  \includegraphics[width=\linewidth]{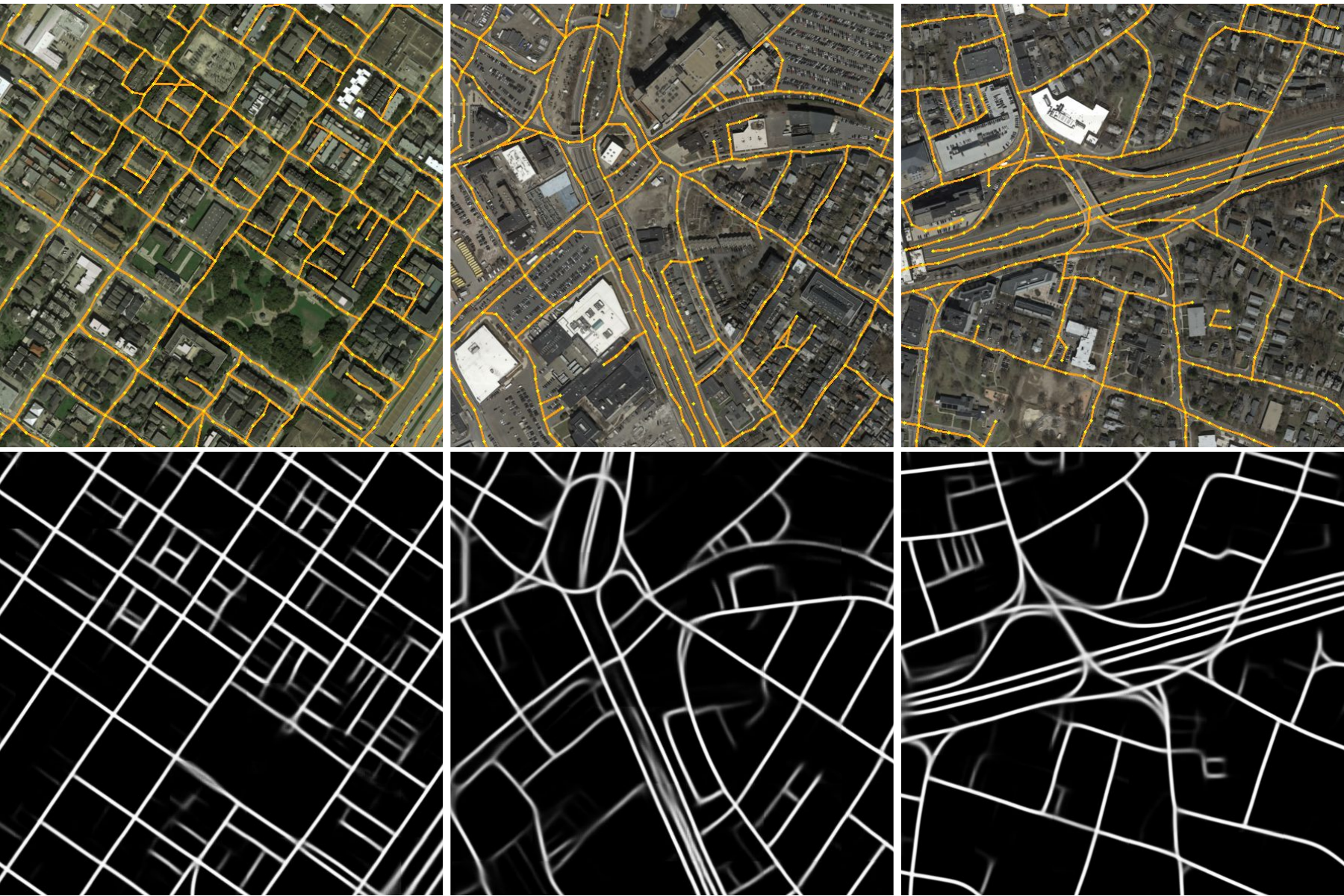}
  \caption{SAM-Road effectively predicts accurate road network graphs for dense urban regions, including roads with complex and irregular shapes, bridges, and multi-lane freeways. The corresponding segmentation masks are sharp and clear.}
  \label{fig:masks_and_graphs}
\end{figure}

Road network graphs are spatial representations of the structure and layout of road networks. They are typically stored in a vectorized format \cite{openstreetmap}, consisting of vertices and edges. The vertices may represent intersections, and edges could stand for road segments. Large-scale road network graphs are vital for various applications: they enable navigation systems like Google Maps to determine optimal routes, assist in path planning for autonomous vehicles \cite{gao2020vectornet,zhao2021tnt}, and help city planners in traffic analysis and optimization \cite{bai2020adaptive}, to name a few. These applications call for accurate and efficient methods to automatically create such graphs, as they require scaling to huge regions and even near-continuous updating \cite{chen2021remote}, which are astoundingly expensive when manually done. Therefore, systems for automatically generating such maps have tremendous application value and are under active research.

Recently, the rapid growth of foundational models \cite{palm,clip,llama,sdiff} showcased their impressive capabilities. These models, which leverage flexible, high-capacity, and scalable architectures such as Transformers \cite{transformer}, are pre-trained through effective self-supervision \cite{mae2022} methods and unprecedentedly large datasets. This endows them with robust semantic reasoning and generalization. Segment Anything Model (SAM) \cite{kirillov2023segment} is such a foundational vision model. Trained with millions of images and billions of masks, it demonstrates unparalleled semantic segmentation capabilities. This raises intriguing questions: How can SAM be applied to the prediction of road network graphs from satellite images, and how good can it be?

In this work, we answer these questions by introducing the SAM-Road model, which adapts the SAM for generating large-scale, vectorized road network graphs. Incorporating domain knowledge from previous research in satellite mapping, we divide the problem into two main components: geometry prediction and topology reasoning. 

We model graph geometry with a set of 2D vertices that, when densely sampled, accurately reflect the graph's overall shape. The SAM-Road model first predicts dense segmentation masks to indicate the likelihood of road elements such as lane segments and intersections, then it employs simple non-maximum suppression to convert the pixels into vertices of the desired density. Leveraging the inherent semantic segmentation capabilities of SAM, this method can effectively capture highly complex shapes (see Figure \ref{fig:masks_and_graphs}), which are common in dense urban areas.

A notable challenge for segmentation-based mapping approaches is the difficulty of inferring topology from dense imagery. This branch of methods often relied on slow, complex and error-prone post-processing heuristics. Inspired by recent advances in graph learning \cite{reltr,relationformer}, we developed a transformer-based graph neural network as the second stage of our model. This network focuses on predicting the local subgraph around each vertex and determining connectivity with nearby vertices to establish the overall graph topology. It utilizes relative vertex positions and image embeddings from the SAM backbone to guide its predictions.

Despite its straightforward design, SAM-Road achieves accuracy comparable to more complex state-of-the-art systems on two widely recognized satellite mapping datasets: City-scale \cite{He2020Sat2GraphRG} and SpaceNet \cite{spacenet2019}. Moreover, for large spatial areas spanning multi-square kilometers, its architecture supports high degrees of parallelism and rapid GPU inference, achieving speeds up to 80 times faster than existing methods. We hope that this work will inspire further exploration of foundational vision models in remote sensing and graph learning tasks.

\section{Related Works}
\label{sec:related_works}

\subsection{SAM and Its Applications}
In 2023, Segment Anything Model \cite{kirillov2023segment} was proposed as a foundational model for image segmentation, showcasing impressive zero-shot and generalization capabilities. Through fine-tuning or direct adoption, SAM has been used in object detection \cite{liu2023grounding}, image inpainting \cite{yu2023inpaint}, segmentation of medical images \cite{ma2023segment, Huang_2024, Wu2023MedicalSA, samed2023}, and remote sensing tasks \cite{RSPrompter}. Existing adaptations of SAM in remote sensing have focused more on simple segmentation and have not yet been applied to the production of road network graphs.


\subsection{Road Network Graph Prediction}

Research on road network graph detection dates back to 2010 \cite{detectroads2010}. Representative methods fall into two categories: segmentation-based and graph-based approaches.

Segmentation-based methods \cite{mattyus2017deeproadmapper, batra2019improved, He2020Sat2GraphRG} treat the task as a dense mask prediction. They represent the road network graph structure through one or more images, each detailing aspects such as road existence, intersections, orientation \cite{batra2019improved}, and connectivity \cite{He2020Sat2GraphRG}. Post-processing heuristics, such as thinning \cite{cheng2017automatic, thinning1984} and path-finding \cite{hdmapnet2022}, are then employed to extract the vectorized graph structure. Benefits of this approach include 1) the ability of segmentation masks to represent complex geometries as a bottom-up volumetric representation \cite{dmtet2021}, and 2) ease of parallel patch-wise inference for large areas, and subsequent result aggregation for refinement. However, the challenge of topology prediction persists: handcrafted heuristics often fail with poor mask quality; even with high-quality masks, deriving topology from them remains ill-formed. There exist no universal heuristics for all complex road structures, like multi-way intersections, multi-lane highways, and overpasses. Moreover, the heuristic tends to rely on CPU-intensive logic, which often becomes the inference speed bottleneck.

Graph-based methods have gained popularity recently, offering a more end-to-end approach. Unlike methods that use intermediate representations like mask images, they directly predict graph nodes and edges in vectorized form. Leading examples include RoadTracer \cite{bastani2018roadtracer}, RNGDet \cite{RNGDet}, and RNGDet++ \cite{xu2023rngdetplus}, with similar advancements in high-definition map generation for autonomous vehicles\cite{mi2021hdmapgen, liu2023vectormapnet, liao2023maptr}. These methods reduce dependence on handcrafted graph generation rules, largely leveraging DETR-like\cite{liu2023vectormapnet, liao2023maptr, zhang2022dino, carion2020endtoend} techniques for geometric element prediction or adopting an autoregressive \cite{bastani2018roadtracer,RNGDet, xu2023rngdetplus, mi2021hdmapgen} approach for incremental graph construction. Despite their strengths and contributions to the state-of-the-art \cite{xu2023rngdetplus}, limitations exist: 1) DETR-like methods struggle with more than a few dozen entities due to the $O(N^2)$ computational complexity of transformer layers, limiting their applicability to city-scale road network graphs with potentially thousands of nodes and edges, and 2) autoregressive methods are difficult to parallelize as they rely on the outcomes of previous steps, significantly slowing down the process.

Our method combines the advantages of segmentation-based and graph-based approaches. It harnesses the exceptional capabilities of SAM to generate a high-quality mask for geometry prediction, and uses a transformer-based graph neural network to directly produce graph structures without handcrafted post-processing heuristics.

\subsection{Graph Representation and Learning}

Graph representation and learning \cite{graphrep2017} involves mapping data to graph structures and applying learning algorithms to understand complex relationships within. Significant advancements have been made in this area with the development of Graph Neural Networks (GNNs) \cite{gnn2018}, Graph Convolutional Networks (GCNs) \cite{gcn2018}, and Transformers adapted for graph data \cite{reltr}. Entities with rich structures can be represented as graphs and predicted by deep nets, such as scene graphs \cite{sceneg2021}, human keypoints \cite{pose3d2024, pose3d2022}, meshes \cite{polygen2020}, and in our case, road networks. The goal is to predict whether a graph edge (road segment) exists between a pair of nodes (vertices). For this type of task, GCN is a suitable architecture choice, as they offer powerful mechanisms for aggregating local subgraph information and understanding node relationships. With multiple layers, long-range dependencies can be captured too. In SAM-Road, we adopt Transformers as a special form of GCN: their self-attention mechanism has a simple form and can automatically select the most relevant context \cite{translation2014, transformer, gao2020vectornet, rqen2018} without any preset structure.


\newcommand{\VertexSet}{$\mathbf{V} \{v_i \in \mathbb{R}^2\}$\,}
\newcommand{\NbrRadius}{$R_\text{nbr}$\,}
\newcommand{\AnEdge}{$(v_i, v_j)$}
\newcommand{\RoadGraph}{$\mathbf{G}$\,}
\newcommand{\EdgeSet}{$\mathbf{E}$\,}
\newcommand{\ImgH}{$H_\text{img}$\,}
\newcommand{\ImgW}{$W_\text{img}$\,}
\newcommand{\FeatH}{$H_\text{feat}$\,}
\newcommand{\FeatW}{$W_\text{feat}$\,}
\newcommand{\FeatD}{$D_\text{feat}$\,}
\newcommand{\VertexIntv}{$d_v$\,}
\newcommand{\MaxNbrs}{$N_\textbf{nbr}$\,}
\newcommand{\MaxSamples}{$N_\textbf{sample}$\,}

\section{Method}

\subsection{Overall Architecture}

\begin{figure*}
  \centering
  \includegraphics[width=\linewidth]{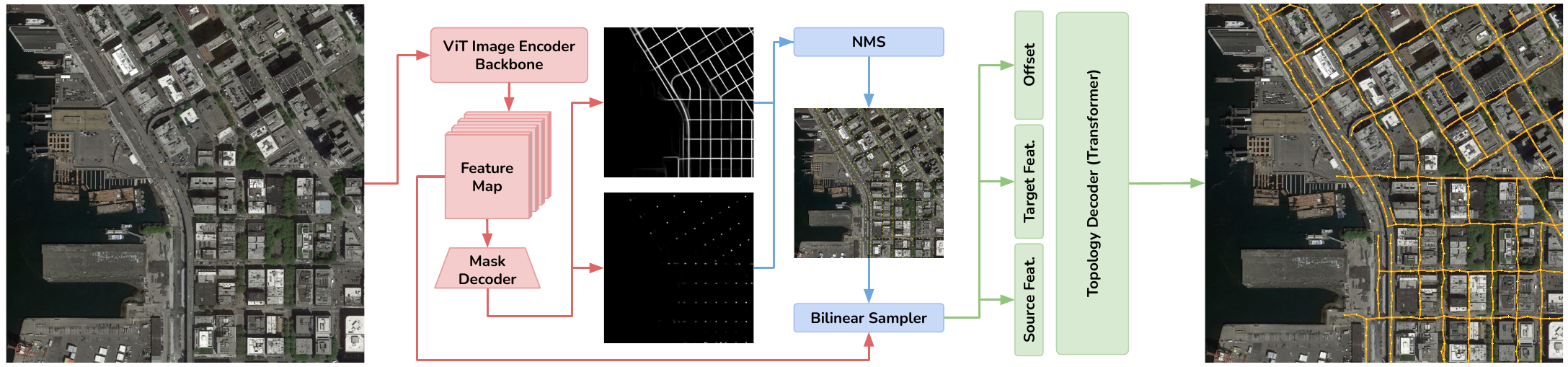}
  \caption{The architecture of our approach, SAM-Road. It contains an image encoder taken from the pre-trained SAM \cite{kirillov2023segment}, a geometry decoder, and a topology decoder. It directly predicts vectorized graph vertices (yellow) and edges (orange) from an input RGB satellite imagery. Better zoom-in and view in color.}
  \label{fig:method}
\end{figure*}

The overall structure of SAM-Road is shown in Figure ~\ref{fig:method}.
It contains an image encoder taken from the pre-trained SAM \cite{kirillov2023segment}, a geometry decoder, and a topology decoder. The model takes as input an RGB satellite imagery. First, the image encoder produces the image feature embeddings. Then, the geometry decoder predicts the per-pixel existence probability, for both roads and intersections. The set of graph vertices \VertexSet 
representing 2D locations is extracted from these masks with a simple non-maximum suppression process, detailed in Algorithm \ref{alg:nms}. Given the predicted vertices, the topology decoder goes over each of them and determines whether it should connect to its nearby vertices within a given radius \NbrRadius, given its local context. For an edge \AnEdge, it predicts the probability that it exists. One edge may be predicted more than once, its final score will be the average. Eventually, the road network graph \RoadGraph is predicted as the sets of vertex $\mathbf{V}$ and edges \EdgeSet.

\subsection{Image Encoder}

The image encoder is taken from a pre-trained Segment Anything Model. We use the smallest ViT-B variant, which has around 80M trainable parameters. It uses a ViT \cite{dosovitskiy2021image} architecture adapted for high-resolution images, as described in ViTDet \cite{vitdet2022}. The image encoder converts an (\ImgH, \ImgW, 3) RGB image into a (\ImgH/16, \ImgW/16, \FeatD) feature map, for the decoders to consume. The image is first divided into 16$\times$16 non-overlapping patches, then each patch is encoded into an embedding vector, producing an (\ImgH/16, \ImgW/16, \FeatD) tensor. A stack of 12 multi-head self-attention layers processes this tensor to the final feature map, alternating between windowed \cite{liu2021swin} and global self-attention. The feature size stays constant along the way. During training, we fine-tune the entire image encoder with $0.1\times$ base learning rate to adapt it to satellite imagery. 

\subsection{Geometry Decoder}

The graph geometry prediction is formulated as a dense semantic segmentation task. There are two main benefits: First, this formulation leverages the extraordinary power of SAM; Second, per-pixel bottom-up representation can handle arbitrarily complex road structures.

The mask decoder has a minimalist design: it's simply 4 transposed convolution layers with $3\times3$ kernels and stride $2$, each doubling the spatial feature resolution and decreasing the channel number. Eventually, it produces two probability maps as an (\ImgH, \ImgW, 2) tensor, with the same size as the input image, representing the existence probability of intersection points and roads. This mask decoder contains about 170K trainable parameters.

After acquiring the masks, the graph vertices are extracted from them. This process converts the dense mask images into a set of sparse vertices, with roughly the same interval \VertexIntv in between. \VertexIntv is selected to be sparse while not hurting geometry accuracy. It's implemented with simple non-maximum suppression: we first drop the pixels under a probability threshold $t$, then traverse them by a descending order of their probability. Pixels within a \VertexIntv radius of the current one are removed. The $(x, y)$ locations of the remaining pixels form the graph vertices \VertexSet. See Algorithm \ref{alg:nms}.


\begin{algorithm}
\caption{Non-Maximum Suppression of Vertices}\label{alg:nms}
\footnotesize
\begin{algorithmic}[1]
\State $\mathbf{V} \gets \emptyset$
\State $t \gets \text{threshold value}$
\State $d_v \gets \text{radius for non-maximum suppression}$
\For{each pixel in the image}
    \If{pixel value $> t$}
        \State Add pixel coordinates $(x,y)$ to $\mathbf{V}$
    \EndIf
\EndFor
\State Sort $\mathbf{V}$ by pixel values in descending order
\For{each $(x, y)$ in $\mathbf{V}$}
    \For{each $(x', y')$ after $(x, y)$}
        \State $d \gets \text{distance between } (x', y') \text{ and } (x, y)$
        \If{$d < d_v$}
            \State Remove $(x', y')$ from $\mathbf{V}$
        \EndIf
    \EndFor
\EndFor

\end{algorithmic}
\end{algorithm}

We predict masks for both intersections and roads for more accurate graph structures at intersections. If only the road mask existed, there would be no guarantee that the center point of an intersection would be kept, producing error patterns like Figure ~\ref{fig:ablation}. To mitigate this: 1) Vertices are extracted from both masks with the same NMS algorithm. 2) The two sets of vertices are joined, with all intersection vertices assigned a higher score than any road vertices. 3) The joined set is then NMS-processed again to produce the final result. This ensures intersection points are kept as much as possible.

\subsection{Topology Decoder}
\label{sec:toponet}

\begin{figure}
  \centering
  \begin{subfigure}{0.49\linewidth}
    \includegraphics[width=\linewidth]{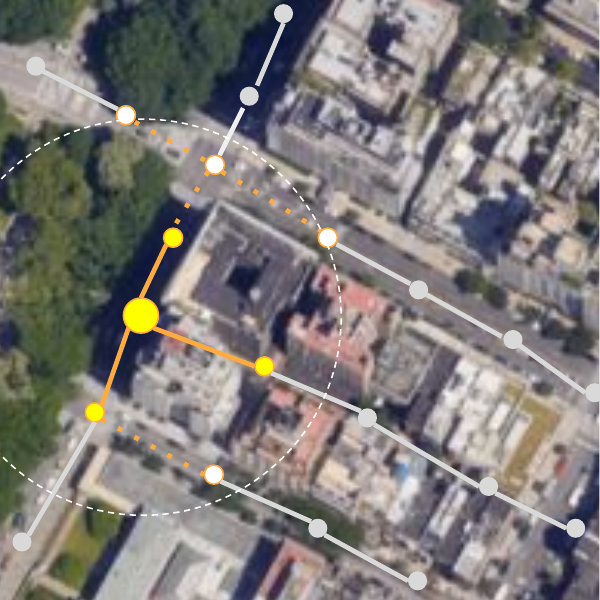}
    \caption{Topology label example}
    \label{fig:topo_label_a}
  \end{subfigure}
  \hfill
  \begin{subfigure}{0.49\linewidth}
    \includegraphics[width=\linewidth]{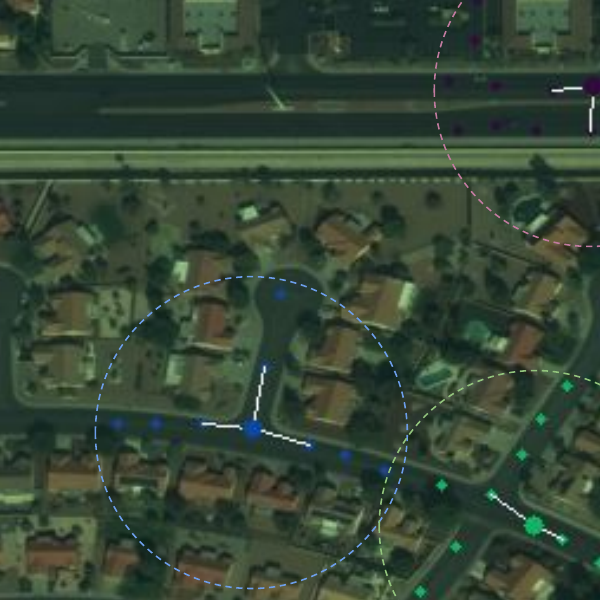}
    \caption{Actual topology samples}
    \label{fig:topo_label_b}
  \end{subfigure}
  \caption{Illustrating the definition of topology labels. In (a), the white dashed circle represents \NbrRadius; the large dot is the source node, and the smaller yellow dots are the target nodes. Orange lines are connected pairs. In (b), a few real topology samples used for training are shown. The query for one source node is shown in the same color. White lines are positive labels and pairs without lines are negative.}
  \label{fig:topo_label}
\end{figure}

The topology decoder "wires up" the predicted graph vertices into the correct structure. It is a transformer-based graph neural network that predicts the existence of edges. It predicts the edge existence probability in small local subgraphs around each vertex. Specifically, for a given source vertex, up to \MaxNbrs nearest vertices are found within a radius of \NbrRadius. These form the target vertices. The topology decoder then predicts whether the source vertex shall connect with each of the targets, based on their spatial layout and image context.

The connection here is defined as "whether two vertices are immediate neighbors on the graph". That is, imagine a breadth-first-search on the road network graph from the source vertex, which stops expanding whenever a) it hits a target vertex or b) the depth (search radius) exceeds \NbrRadius - a target vertex is only connected to the source if it is visited by the search. This is further illustrated in Figure ~\ref{fig:topo_label}.

We formulate the topology prediction task as a binary classification problem on the $(v_\text{src}, v_\text{tgt})$ vertex pairs, conditioned on the image context. The input of the decoder is a sequence of high-dimensional feature vectors $\{ (f^\text{src}, f^\text{tgt}_{k}, \Vec{d}_k) \mid 0 \leq k < N_\text{nbr}\}$\, where $f^\text{src}$ and $f^\text{tgt}_{k}$ are the vertex features. They are image embedding vectors acquired by bilinear sampling from the SAM image feature map at the source and target vertex locations. $\Vec{d}_k$ is the offset from the source to the $k$-th target, encoding the relative spatial layout of the vertices of interest. These vectors are concatenated to a tensor shaped (\MaxNbrs, $2D_\text{feat} + 2$), then projected to a (\MaxNbrs, \FeatD) feature tensor. We treat the \MaxNbrs dimension as sequence length and pass it through 3 multi-head self-attention layers with ReLU activations for message-passing to understand the multi-hop structures. The interacted feature sequence shaped (\MaxNbrs, \FeatD) is fed into a linear layer to get the \MaxNbrs binary classification logits. A sigmoid layer turns these into (0, 1) probabilities, indicating how likely the edge exists.

\subsection{Label Generation}

\textit{Mask Labels.} For road mask labels, we rasterize the ground-truth road lines, by drawing each edge as a line segment, with a width of 3 pixels. The pixels covered by the line segments are set to 1, and others are 0. For intersection labels, we find all the graph vertices with a degree not equal to 2 and render them as circles with a radius of 3 pixels. This is partially inspired by the OpenPose \cite{openpose} work which represents human keypoint graphs as heatmaps.

\textit{Topology Labels.} During training, we don't run the vertex extraction process. The topology decoder is trained in a teacher-forcing \cite{teacherforcing1989} manner, where the vertices being asked are not from model prediction, but sampled from ground-truth road network graphs to emulate the predictions. This is done by first subdividing the ground-truth graph and then running the same NMS procedure as the inference stage. To emulate various NMS results, a uniform random score is assigned to each subdivision vertex. 

Having the emulated vertex predictions, we randomly sample \MaxSamples source vertices and apply the rules described in section \ref{sec:toponet} to find its targets and connectivity labels. Further, a small random Gaussian perturbation is applied to the vertex coordinates to emulate the prediction noise for better generalization.

The satellite imagery in the datasets used covers large square areas up to 4 square kilometers \cite{He2020Sat2GraphRG}, therefore we randomly crop the RGB image, ground-truth masks, and graphs into smaller patches to get more training samples and keep memory consumption manageable.

\subsection{Sliding-window Inference for Large Regions}
\label{sec:sliding_window}
\begin{figure}
  \centering
  \includegraphics[width=\linewidth]{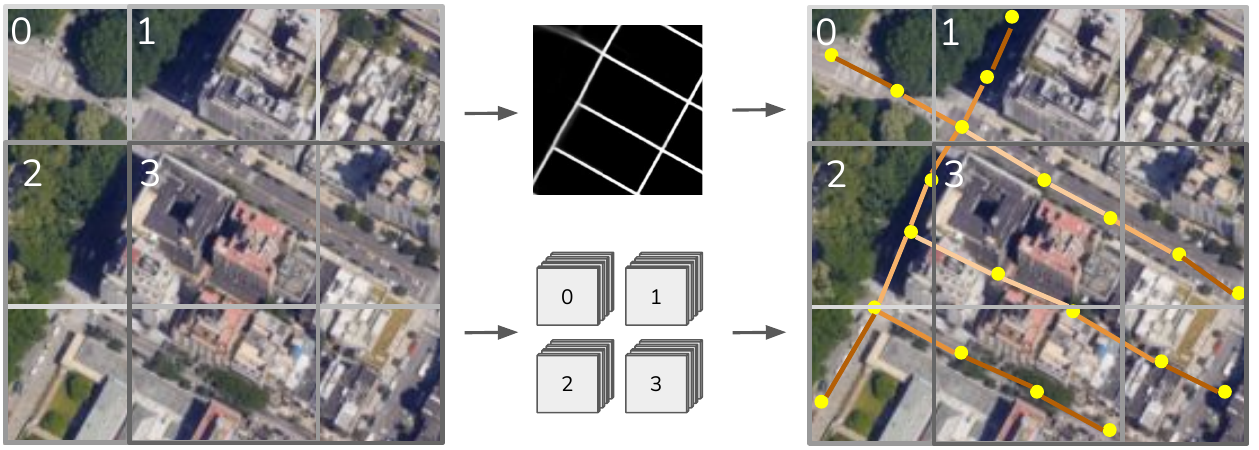}
  \caption{SAM-Road can predict the entire road network graph for arbitrarily large regions by operating in a sliding-window manner. 0-3 represent 4 overlapping windows. It first extracts the global nodes, caches the per-window embeddings, and then aggregates the per-window edge predictions.}
  \label{fig:parallel_infer}
\end{figure}

SAM-Road can predict the entire road network graph for arbitrarily large regions by operating in a sliding-window manner, as shown in Figure ~\ref{fig:parallel_infer}. The predictions within each window can be aggregated to improve accuracy. Fusing multiple observations is a common practice in vision applications \cite{hdmapnet2022, offboard3d2021, congrui2017impression, vistemp2021} to effectively suppress noise. For SAM-Road, this applies to both geometry and topology.

For geometry, the per-window masks are fused to a large mask before vertex extraction, where each pixel value is the sum of all observed probabilities divided by the time it is observed. The NMS process is run on the fused global mask to get the global graph vertices.

For topology, when it comes to large regions, the topology decoder is run in a second pass after extracting the global vertices. The per-window image feature maps are cached, and for each window, the topology decoder infers the graph edges for the global vertices within that window, based on its image feature map. Since the vertices here are global, each edge prediction within each window can vote towards an edge in the global graph. The final edge probability in the global graph is the average of all observations similar to the mask.

It's also worth noting that the per-window inferences are completely independent of each other and can be done fully in parallel. This enables SAM-Road to be significantly faster (See Table ~\ref{tab:speed_comparison}) than the state-of-the-art RNGDet++ \cite{xu2023rngdetplus} that reconstructs the graph in an auto-regressive manner. The ease of multi-window aggregation for quality refinement, akin to dense semantic segmentation, is also uncommon for typical graph-based methods. SAM-Road can flexibly trade-off between speed and accuracy, by varying the stride size in sliding-window inference, as shown in Table ~\ref{tab:speed_accuracy}.

\section{Experiments}

\subsection{Datasets}
We conduct our experiments on two datasets: City-scale \cite{He2020Sat2GraphRG} and SpaceNet \cite{spacenet2019}. The City-scale dataset includes 180 satellite images of 20 U.S. cities, each image has $2048 \times 2048$ pixels, and 29 are for testing. The SpaceNet dataset contains 2549 images of $400 \times 400$ pixels of cities around the world including Shanghai, Las Vegas, and more. 382 of them are for testing.

Both datasets have a 1 meter/pixel resolution. The ground-truth vector graphs of the road network are supplied. The two datasets feature diverse environments and road network patterns, facilitating conclusive experiments.

\subsection{Metrics}
We employ TOPO \cite{topometric}, an evaluation metric tailored for road network graphs. TOPO randomly samples candidate vertices in the ground truth and finds its correspondence in the prediction. It then compares the similarity of reachable sub-graphs from the same vertex of the two graphs in terms of precision, recall, and F1. It focuses on geometric accuracy with a heavy penalty for incorrect disconnections. 

We also utilize APLS (Average Path Length Similarity) \cite{spacenet2019} to evaluate the topological correctness. For a random vertex pair $(v_1, v_2)$ on the ground truth and their correspondences in the prediction $(\hat{v_1},\hat{v_2})$, we evaluate the model by comparing the shortest distance between $(v_1, v_2)$ and between $(\hat{v_1},\hat{v_2})$. Smaller distance difference indicates high topological similarity. 

\subsection{Implementation Details}

For both datasets, $d_v$ is 16 pixels (meters), $R_\text{nbr}$ is 64 pixels (meters), $N_\text{nbr}$ is 16, $D_\text{feat}$ is 128. At training time, For City-scale, we sample image patches of $512\times512$ pixel, the batch size is 16 and we sample 512 source points for topology query per image patch. For SpaceNet, the batch size is 64 due to using image patches $256\times256$ pixel. We sample 128 source points per patch. When there are fewer than $N_\text{nbr}$ available target nodes to query, we use attention masking to ensure the interaction only happens between the valid vertices. 

We applied simple augmentations to boost data diversity. 1) Rotational: we randomly rotate the patch by the multiple of 90 degrees. 2) Translational: different from previous works that usually pre-crops the patches by a fixed grid and stores them to disk, we load the entire dataset in memory, and randomly sample patches in continuous spatial coordinates. This can be seen as a random-translation augmentation.

Masks and topology prediction are essentially binary classifications. We use the vanilla binary cross entropy loss for all of them and don't apply any loss re-weighting in this work. We take the mean loss of all valid entries. The three sub-tasks have equal loss weight, and the total loss is just adding them together. 

We use the Adam optimizer with base LR of 0.001, which applies to the randomly initialized mask decoder and topology decoder. We use the default weight initialization of PyTorch. The image encoder is fine-tuned with $0.1\times$ base LR. LR is constant during training, with no scheduling tricks applied. We train SAM-Road on the two datasets respectively till validation metrics plateaus.

At inference time, we use 16x16 sliding window inference for the main results. To determine the threshold for the binary classifiers (intersection, road, edge connection), we find the threshold that gives the highest F1 score on the validation set. Note that this is just for isolating away the effect of threshold choice in the experiments, and is not critical for SAM-Road performance, as evidenced by the result that just uses 0.5 for everything in Table ~\ref{tab:ablation} (A vs H).

All experiments are conducted on one RTX 4090 GPU.

\subsection{Evaluating Road Network Prediction}

\begin{figure*}
  \centering
  \includegraphics[width=0.908\linewidth]{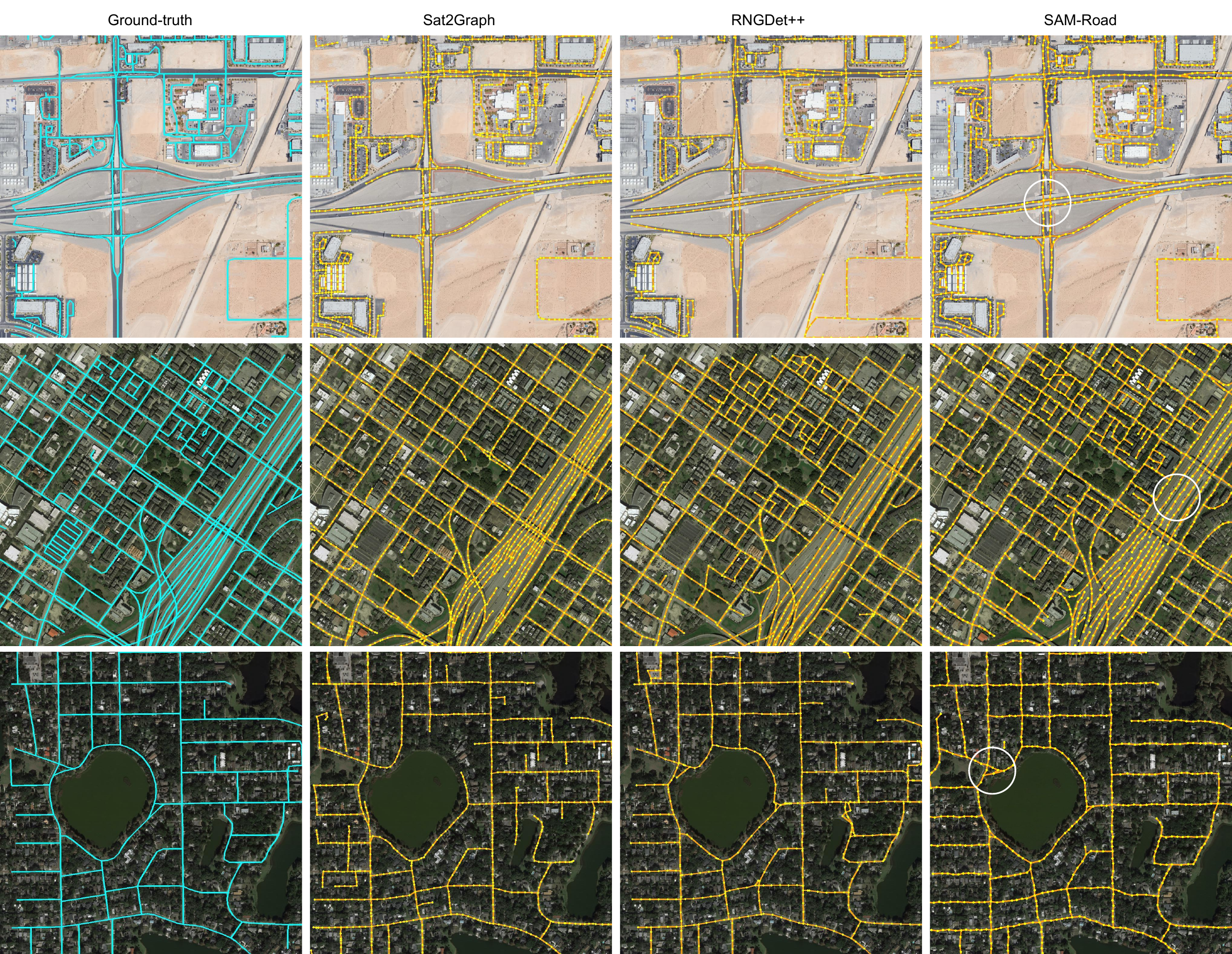}
  \caption{The visualized road network graph predictions of SAM-Road and two baseline methods. Better zoom-in and view in color. Overall, SAM-Road generates highly accurate predictions. The circles highlight especially challenging spots: in the first area, SAM-Road correctly predicts the overpass structure. In the second one, SAM-Road gives superior results for the parallel freeways. The third spot shows an irregular intersection where the two baselines fail.}
  \label{fig:compare}
\end{figure*}

\begin{table*}[ht]
\footnotesize
\centering
\begin{tabular}{@{}lcccccccc@{}}
\toprule
& \multicolumn{4}{c}{City-scale Dataset} & \multicolumn{4}{c}{SpaceNet Dataset} \\
\cmidrule(lr){2-5} \cmidrule(lr){6-9}
Methods & Prec.$\uparrow$ & Rec.$\uparrow$ & F1$\uparrow$ & APLS$\uparrow$ & Prec.$\uparrow$ & Rec.$\uparrow$ & F1$\uparrow$ & APLS$\uparrow$ \\
\midrule
Seg-UNet \cite{unet2015} & 75.34 & 65.99 & 70.36 & 52.50 & 68.96& 66.32 & 67.61 & 53.77 \\
Seg-DRM \cite{mattyus2017deeproadmapper} & 76.54 & 71.25 & 73.80 & 54.32 & 82.79 & 72.56 & 77.34 & 62.26 \\
Seg-Improved \cite{batra2019improved} & 75.83 & 68.90 & 72.20 & 55.34 &81.56 & 71.38 & 76.13 & 58.82\\
Seg-DLA \cite{dla2018} & 75.59 & 72.26 & 73.89 & 57.22 & 78.99& 69.80& 74.11 &56.36\\
RoadTracer \cite{bastani2018roadtracer} & 78.00 & 57.44 & 66.16 & 57.29 & 78.61 & 62.45 & 69.90 & 56.03\\
Sat2Graph \cite{He2020Sat2GraphRG} & 80.70 & 72.28 & 76.26 & 63.14 & 85.93& 76.55 &80.97&64.43\\
TD-Road \cite{tdroad2022} & 81.94 & 71.63 & 76.43 & 65.74 & 84.81& \textbf{77.80} &81.15&65.15\\
RNGDet \cite{RNGDet} & 85.97 & 69.78 & 76.87 & 65.75 &90.91 &73.25
&81.13 &65.61\\
RNGDet++ \cite{xu2023rngdetplus} & 85.65 & \textbf{72.58} & \textbf{78.44} &67.76 &91.34& 75.24 & \textbf{82.51} &67.73\\
\midrule
SAM-Road & \textbf{90.47} & 67.69 & 77.23 & \textbf{68.37} & \textbf{93.03} & 70.97 & 80.52 & \textbf{71.64} \\
\bottomrule
\end{tabular}

\caption{Comparison with existing methods on different datasets. SAM-Road achieved the highest TOPO precision of $90.47\%$ on City-scale and $93.03\%$ on SpaceNet. It also shows the highest APLS metric of on both sets. Overall the graph accuracy is among the very top. SAM-Road leans more towards precision in TOPO metrics, this might be due to the low positive / negative example ratio in its binary classification tasks.}
\label{tab:main_numbers}
\end{table*}

\begin{figure}
  \centering
  \includegraphics[width=\linewidth]{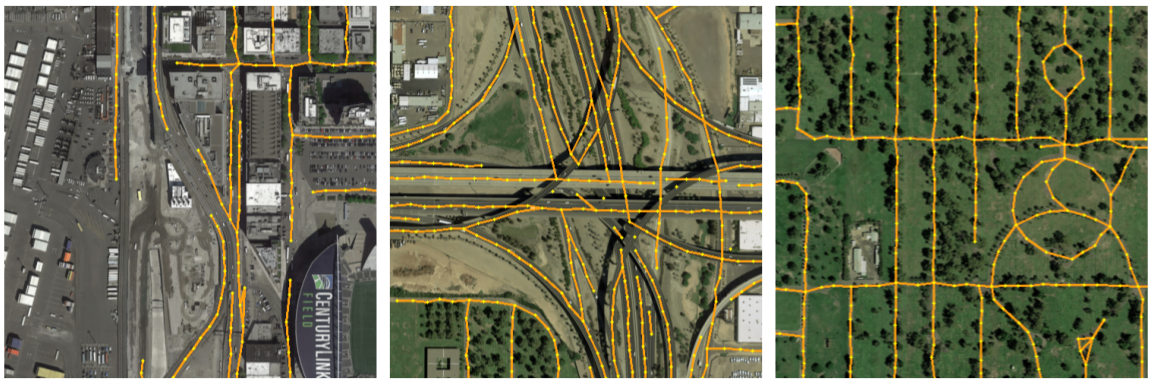}
  \caption{Some error patterns. Left: geometry decoder missed the road segment in the middle. Middle: topology decoder missed connections in a complex interchange. Right: an interesting case where SAM-road predicts the trails in a park which are not part of the label. }
  \label{fig:failure_cases}
\end{figure}

Qualitative results of SAM-Road predicting large-scale road network graphs can be found in Figure ~\ref{fig:compare}. The results are shown side-by-side with two baselines and the ground-truths. Some error examples can be found in Figure \ref{fig:failure_cases}. Overall, SAM-Road predicts highly accurate road networks even under very challenging circumstances, e.g. many blocks and intersections in dense urban areas, curvy roads with irregular shapes, overpasses, and multi-lane highways.

We benchmark SAM-Road on City-scale and SpaceNet benchmarks against other methods, quantitative results are shown in Table \ref{tab:main_numbers}. We compare several baselines, including segmentation-based (Seg-UNet, Seg-DRM, Seg-Improved, Seg-DLA, Sat2Graph) and graph-based (RoadTracer, RNGDet, RNGDet++). The TOPO metric, which evaluates local graph structure similarity, is on par with state-of-the-art, RNGDet++, despite that SAM-Road has a much simpler structure. The APLS metric of SAM-Road achieves a new state-of-the-art. APLS captures long-range topological and geometrical structure - this indicates the effectiveness of our transformer-based topology decoder and graph representation. 

Such performance should largely be attributed to SAM, the powerful foundational vision model. As shown in Figure \ref{fig:masks_and_graphs}, the predicted masks are sharp and clear, enabling precise geometry prediction. The SAM image features are also informative vertex embeddings containing rich semantic meanings, as evident in the accurate topology predictions.


\subsection{Speed and Accuracy Trade-off}

\begin{table}
  \centering
  \footnotesize
  \begin{tabular}{@{}lcc@{}}
    \toprule
    Method & City-scale Dataset & SpaceNet Dataset \\
    \midrule
    Sat2Graph & 150.6 min & 69.0 min \\
    RNGDet++ & 231.0 min & 112.8 min \\
    SAM-Road & \textbf{4.6 min} & \textbf{8.2 min} \\
    \bottomrule
  \end{tabular}
  \caption{The inference time for the three methods, on both City-scale and SpaceNet datasets. Ours is within 10 minutes while the other two methods take 1-2 hours.}
  \label{tab:speed_comparison}
\end{table}

\begin{table}[ht]
\centering
\resizebox{\columnwidth}{!}{
\begin{tabular}{@{}lcccccc@{}}
\toprule
& \multicolumn{3}{c}{City-scale Dataset} & \multicolumn{3}{c}{SpaceNet Dataset} \\
\cmidrule(lr){2-4} \cmidrule(lr){5-7}
Setup & Time Cost & F1$\uparrow$ & APLS$\uparrow$ & Time Cost & F1$\uparrow$ & APLS$\uparrow$ \\
\midrule
$16\times16$ & 4.6 min & 77.23 & 68.37 &  8.2 min & 80.52 & 71.64  \\
$8\times8$ & 3.3 min & 77.20 & 67.21 &  3.1 min & 80.84 & 71.12  \\
$4\times4$ & 2.9 min & 77.00 & 67.03 &  1.7 min & 80.85 & 70.88  \\
\bottomrule
\end{tabular}
}
\caption{The time cost with different stride sizes in sliding-window inference, on both datasets.}
\label{tab:speed_accuracy}
\end{table}

SAM-Road is also highly efficient, thanks to its parallelized inference and that it doesn't require complex CPU-heavy post-processing heuristics. We measure the inference time to produce the complete graphs for the test sets of both datasets. The main results use $16\times16$ windows and are already $40\times$ faster than RNGDet++ on the City-scale dataset, and $10\times$ faster on the SpaceNet dataset, as shown in Table \ref{tab:speed_comparison}. As mentioned in Section \ref{sec:sliding_window}, SAM-Road can trade accuracy for more speed by sparsifying the sliding windows. Table \ref{tab:speed_accuracy} shows the result such trade-off. Using fewer windows can further provide $2\times$ to $4\times$ speed-up, with a minor accuracy drop.

\subsection{Ablation Studies}

\begin{table}[ht]
\centering
\resizebox{\columnwidth}{!}{
\begin{tabular}{@{}c|cccccc|cc@{}}
\toprule
Variant & Opt & SAM & TFM & Offset & F-target & Itsc & F1$\uparrow$ & APLS$\uparrow$ \\
\midrule
A & \checkmark & \checkmark & \checkmark & \checkmark & \checkmark & \checkmark & 77.23 & 68.37  \\
B & \checkmark &    & \checkmark & \checkmark & \checkmark & \checkmark & 31.79 & 12.39  \\
C & \checkmark &  \checkmark  &    & \checkmark & \checkmark & \checkmark & 73.75 & 59.39  \\
D & \checkmark &  \checkmark  &  \checkmark  &    & \checkmark & \checkmark & 77.36 & 66.67  \\
E & \checkmark &  \checkmark  &  \checkmark  & \checkmark &   & \checkmark & 77.42 & 67.08  \\
F & \checkmark &  \checkmark  &  \checkmark  & \checkmark & \checkmark &  & 71.94 & 64.62  \\
G & \checkmark &  \checkmark  &    &  &  & \checkmark & 69.21 & 63.32  \\
H &  &  \checkmark  &  \checkmark  & \checkmark & \checkmark & \checkmark & 76.05 & 67.95  \\
\bottomrule
\end{tabular}
}
\caption{The SAM-Road variants compared for ablation studies. Opt: using optimized score thresholds. SAM: using pre-trained SAM. TFM: using a transformer for topology prediction. Offset: taking relative offsets in topology decoder. F-target: topology decoder takes target node feature. Itsc: predict intersection masks.}
\label{tab:ablation}
\end{table}

\begin{figure}
  \centering
  \includegraphics[width=\linewidth]{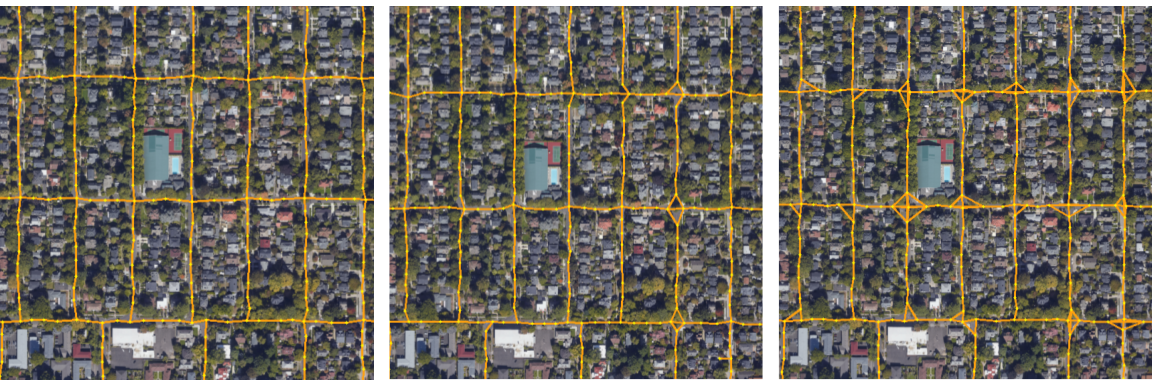}
  \caption{Left: standard SAM-Road. Middle: no intersection mask. The intersections are noticeably noisier. Right: using an A-star algorithm for topology prediction, which induces many false positive connections. }
  \label{fig:ablation}
\end{figure}

We conduct ablation experiments to study the effects of the key design choices on the City-scale dataset. The results are shown in Table \ref{tab:ablation}. 

 \textbf{How important is using the pre-trained SAM model?} A vs B proves it is critical. We repeated the experiment with the same ViT-B architecture with only ImageNet1K and MAE pre-training \cite{vitdet2022}, and the results were far worse. This is not surprising, as City-scale and SpaceNet datasets are quite small in this era, especially when using large patch sizes (E.g. 512), resembling few-shot learning. The large-scale pre-training on datasets like SA-1B used by SAM seems critical for the generalization capability. Maybe it's due to this reason that the baseline methods have to rely on smaller patches for more training examples and adopt weaker backbones with more inductive bias like CNNs.

We also studied the importance of the topology decoder's design choices.

\textbf{Whether using a transformer.} A vs C: we tried removing it and simply connecting a dense layer directly to the pair features. This makes the query unaware of other targets. Both geometry and topology performance drops. This is understandable: all nodes in the subgraph being asked shall be visible to the net, otherwise, there are ambiguities about whether two nodes shall connect given the definition in Section \ref{sec:toponet}.

\textbf{Whether taking the vertex offsets as input.} A vs D shows a slight performance drop. Without the offset, the topology decoder no longer has a clear view of the local geometrical layout, which may hinder the topology reasoning and cause false-positive connections and discontinuities.

\textbf{Whether taking the target vertex feature as input.} A vs E shows a minor performance drop. Interestingly, not using the target node features doesn't harm performance too much. This might be because ViT-B has a sufficiently large effective field of view due to the transformer architecture, and the source feature alone contains sufficient image context in the region.

\textbf{Whether using the learning-based topology decoder.} A vs G shows that it's critical for SAM-Road's performance. Intuitively, a naive method that might achieve a similar effect is just to run a pathfinding algorithm between a pair of vertices, using the road existence map as the cost field, and see if there's a sufficiently low-cost path between the two without passing through other vertices. We implemented such a variant G using an A-star algorithm. Metrics are much worse, as qualitatively shown in Figure ~\ref{fig:ablation}. This approach can mess up intersections, overpasses, and close parallel roads.

\textbf{Whether predicting the intersection vertices.} This is answered by A vs F. Predicting intersection points is important for building correct intersection structures as shown in ~\ref{fig:ablation}. Without it, both metrics drop.
\section{Limitations and Future Work}

One current limitation of SAM-Road is we have not designed specific approaches to more accurately handle overpasses. There is an ambiguity for the topology decoder at the exact point where overpassing roads intersect, as the correct answer depends on which layer is being asked. This issue is minor though, as most vertices are not at these spots. Future work could improve this by predicting an overpass heatmap to suppress vertex formation at these locations.

In addition, in this work, we only used the smallest Segment Anything model, ViT-B. Larger variants may be explored as a future work, where we hope to explore parameter-efficient tuning methods, such as LoRA \cite{hu2021lora}.

We are also interested in exploring the integration of other state-of-the-art foundational models, such as DINOv2 \cite{dinov2_2023}, PaLI \cite{chen2023pali} and GPT-4V \cite{gpt42023} with graph learning.

\section{Conclusion}
We demonstrate the power of SAM \cite{kirillov2023segment}, a foundational vision model on a graph learning task. It reaches state-of-the-art accuracy with a simple design while being much more efficient. This indicates a high-capacity model with massive pre-training can be a strong graph representation learner.

\nocite{thibaux2023stop}
\nocite{unlock2023}
\nocite{slidingbert}
\nocite{xu2020estimating}
\nocite{pathfinding1}
\nocite{pathfinding2}
{
    \small
    \bibliographystyle{ieeenat_fullname}
    \bibliography{main}
}


\end{document}